# A Dynamic Approach to Rhythm in Language: Toward a Temporal Phonology

Robert Port, Fred Cummins and Michael Gasser
Linguistics and Cognitive Science
Indiana University

August 7, 1995

## Abstract

It is proposed that the theory of dynamical systems offers appropriate tools to model many phonological aspects of both speech production and perception. A dynamic account of speech rhythm is shown to be useful for description of both Japanese mora timing and English timing in a phrase repetition task. This orientation contrasts fundamentally with the more familiar symbolic approach to phonology, in which time is modeled only with sequentially arrayed symbols. It is proposed that an adaptive oscillator offers a useful model for perceptual entrainment (or 'locking in') to the temporal patterns of speech production. This helps to explain why speech is often perceived to be more regular than experimental measurements seem to justify. Because dynamic models deal with real time, they also help us understand how languages can differ in their temporal detail—contributing to foreign accents, for example. The fact that languages differ greatly in their temporal detail suggests that these effects are not mere motor universals, but that dynamical models are intrinsic components of the phonological characterization of language.

## 1 The Description of Language

### 1.1 Dynamic and Symbolic Models of Language

By long tradition, human language has been described in terms of strings of discrete, formal symbols (Saussure, 1916; Bloomfield, 1933; Chomsky and Halle, 1968). The basic units of each of the components of a grammar, from syntax through morphology and phonology, are discrete, static information packets arranged in sequential strings. It is assumed that these are subsequently read out during language production and physically implemented as speech. The phonetic component of the speech production system is supposed to take the symbolic units of phonology as input and produce movements by the speech articulators (Chomsky-Halle, 1968). This mapping from the mental and symbolic to continuous-time physical events is quite curious and problematic (Fowler et al., 1981) because the physical phonetic events of speech are not fundamentally different from other kinds of physical events—the motion of a pendulum or the act of taking a step—yet cognitive symbols or 'mental objects' are very different in the kind of time scale that they assume. As conceptualized by linguists, the symbols of natural language exist in some idealized spacetime, one in which events like rule application take place instantaneously and time itself is modelled only by the serial positioning in some buffer of symbols that are themselves timeless.

In describing all classes of physical events, from astronomy to mechanics to neuroscience, the natural sciences conventionally employ descriptions based on dynamical systems theory. Since speech gestures, words and sentences are also physical, continuous events observable in speech gestures, we may propose as a working assumption that *linguistic units are events in time*. From this (linguistically radical) point of view, time and the temporal structure of linguistic events, at all levels from phonetic to syntactic and semantic become central problems. Through development of understanding of 'peripheral' temporal structure, that is, the temporal structure in the speech stimulus, we may hope to find keys to understanding higher-level cognitive aspects of language and what their temporal structure might be. Sentences and conversations are produced in time and interpreted in time. But if all levels of language take place in time, then the cognitive and the physiological must be time-locked to each other. The curious mapping that tries to build a bridge from static symbols to continuous-time physiology dissolves away. Both symbols and speech gestures might embody similar kinds of dynamic structure.

The mathematics appropriate for describing temporal events is *dynamic systems theory*.[1] In order to deal with time in language, it is necessary to approach language from a vantage point that is not familiar to linguists, by using the conceptual tools of dynamic systems theory to account for entities normally given symbolic description.

---

[1] There are now a number of approachable introductions to dynamic systems. An intuitive approach is provided by Abraham and Shaw (1983). For an introduction to the application of dynamic systems theory to problems in cognitive science, see Norton's mathematical introduction (Norton, 1995) to Port and van Gelder (1995). Kelso et al.(1986) give an interesting application of dynamic principles to speech.

The temporal structure of speech production is an obvious choice as a problem for exploration using dynamics. Within the tradition of symbolic description, these temporal phenomena are addressed under the rubrics of 'stress' and 'meter' (Liberman, 1978; Liberman and Prince, 1977; Hayes, 1985) and are modelled using hierarchically arrayed but internally static symbols. We suspect that many of these linguistic phenomena can be accounted for naturally as temporal phenomena. Only relatively recently have research strategies for investigating the temporal structure of linguistic units gradually been developed (Vatikiotis-Bateson, 1993; Browman and Goldstein, 1992; Saltzman and Munhall, 1989; Anderson et al., 1988; Petitot, 1988; Kelso et al., 1986).

Dynamic models have been applied with striking success to many kinds of data regarding motor behavior. Work by many researchers (including Bernstein, Turvey, Kelso, Saltzman, Grossberg and many others) has shown that dynamic systems theory in one form or another offers a plausible framework for describing the coordination of motor activity. Obviously speech too is susceptible to such analysis. But how relevant is this work to theories of language? Can this have any relevance to linguistics? We think so, but we acknowledge that any success of nonlinear dynamic models for motor control offers only weak evidence regarding something as abstract as a grammar. After all, while motor control is in part a cognitive process, it still always involves motion by real physical objects (e.g., the tongue and jaw). So linguists may feel justified in dismissing the theory of motor control as irrelevant for a theory of language.

However, speech perception is a purely cognitive process and does not involve the macroscopic motion of massive structures. What if it could be shown that the process of speech perception for prosodic structures is appropriately modeled by an abstract dynamic system—one that resembles the dynamic structures employed in motor control for skilled actions? We may imagine a system that acts (in appropriate situations) as though it had mass and stiffness. Such a system would be quite abstract, even though it runs in time. When it is modeling stimulus patterns that reflect the motion of physical objects (e.g., jaws, lips and vocal cords), the system should act as though it had masses and springs. The goal of our research program is to find out what kind of dynamic system could model both perception and production of linguistically controlled speech gestures. Since we focus primarily on the perception problem, there are usually no physical masses involved. Some region of neural tissue is responsible for implementing these dynamics. Since this perceptual dynamic is learned differently for each language, it would form the basis of many aspects of foreign accent (Tajima et al., 1994). The conclusion we endorse is that if speech production works on dynamic principles and speech perception can be shown to do so as well, then it becomes very plausible that the phonology of any language is itself simply a particular complex dynamic system—a system based on non-linear dynamic systems such as oscillators which exist at levels like the syllable and foot.

In this paper, we will look at evidence for speech rhythm, where rhythm is defined by the repetition of similar events after similar time intervals. We will need to study both the temporal structure of physical speech signals and also develop a psychological model which can perform the right kind of measurements on the physical signal and make the appropriate predictions (about the similar events and similar intervals). We need to measure linguistically relevant durations: voice-onset time, vowel durations, interstress intervals etc. (Port, 1986; Port, 1990; Port et al., 1995). These measurements are language-specific, and so we anticipate that a general perceptual model for metrical systems will be individually parameterized for each language.

We are hopeful that eventually many of the phenomena currently described using traditional linguistic symbolic concepts such as 'stress', 'metrical hierarchy', 'foot', 'short vowel', 'stress-timing', etc. can be reinterpreted as manifestations of a general perceptual model. A model of this kind would incorporate continuous-time dynamic notions like oscillator, velocity, phase angle, toroidal state spaces, stiffness and entrainment, in order to deal with the phonological units currently interpreted as hierarchically structured symbols. If this research program is successful, it will provide for each language a linguistic description which, unlike standard symbolic models of meter, will require no additional clocking devices to perceive speech or produce it in real time since the model will already live in time.

Despite the radical nature of these proposals, we think it is quite appropriate to characterize our specific research projects below in conventional terms as a search for universal properties of the temporal structure of speech. From this point of view, the project is analogous to the standard 'search for linguistic universals' that serves as the rationale for research by many modern linguists. For example, it is likely that temporal alternations of strong and weak elements in speech production are widespread across languages— possibly even universal (as proposed by Liberman and Prince, 1977). It also appears likely that periodicity or near periodicity at some temporal level or another is a universal of language (as was claimed by Abercrombie, 1967 and hinted at by Pike, 1945). We seek general dynamic mechanisms for rhythm and meter, suitable for any language (and probably for music as well). This paper reports some work which makes a small start toward uncovering a subdiscipline within phonology that might be called the *temporal phonology* of language. This area is concerned with the perception and production of speech in time, and with the description of natural languages using dynamic systems.

## 1.2  Rhythm in Speech

Poets, phoneticians and most ordinary speakers share the intuition that speech is often rhythmically performed. As far as we know, all linguistic communities exhibit some overtly rhythmical styles of speech—whether it is characterized as poetry, song, chant, preaching, some style of public declamation or merely worksong. Many communities also have conventional forms of group recitation or responsive reading where a text is recited in unison (e.g., the American "Pledge of Allegiance"). Given the apparent appeal of such genres of speech, one might look for similarities between these styles and normal spoken language (as proposed by Abercrombie, 1967) in order to reveal the structure of language itself.

But among linguists, there remain longstanding arguments about the extent to which perceived rhythm in normal prose reflects quantitative timing constraints or whether the perception is an experience based merely on structured alternations of serially ordered units of different strengths (Liberman, 1978; Boomsliter and Creel, 1977). Yet even ordinary prose speech is sometimes described in rhythmical terms, even to the extent of representation by musical notation (Jones, 1932; Martin, 1972).

Kenneth Pike (1945) characterized some languages (like English and Russian) as "stress-timed," suggesting that in these languages the "time interval between the beginning of prominent syllables is somewhat uniform" (p. 34) and their component intervals (like syllables and segments) were stretched or compressed to make the onsets of stressed syllables more equally spaced. He claimed that some other languages (like French and Spanish) were "syllable-timed," meaning that each syllable is produced with an equally spaced psychological *beat* of its own (p. 35). Abercrombie asserted boldly that "all human speech possesses rhythm" and further claimed that all languages in the world are either stress-timed or syllable-timed (Abercrombie, 1967). He proposed that listeners who speak the former rhythmic type of language would have "expectations about the regularity of succession" of stresses and the latter type would have expectations about syllables.

But the relationship between such perceptual structures and quantitative real-time measurements has remained controversial among linguists. This may be due in part to the preference for symbolic description of linguistics, but results primarily from the failure of most claims of temporal regularity to survive careful physical measurement. After all, the simplest way to translate discrete linguistic symbols into hypotheses about temporal extent, is to predict perfect isochrony. Since the symbols are all the same size, their manifestations in time might also be expected to be the same size, presumably down to the level of 'motor noise.' Of course, finding evidence for perfect isochrony (i.e., equal temporal spacing) is not very likely. As long ago as 1939, Classé used the earliest kymographic methods of measurement (based on simple levers attached to an articulator at one end and a writing instrument at the other) and looked for evidence of isochronous intervals between stressed syllable onsets in English. Given his highly irregular results, he suggested that interstress isochrony may be only an "underlying tendency" (Classé, 1939). Other attempts (Shen and Peterson, 1962; Bolinger, 1965; Uldall, 1971; Couper-Kuhlen, 1993) have also produced results that supported much weaker claims than isochrony. Evidence of isochrony in so-called syllable timing languages is also typically discouraging (Wenk and Wioland, 1982; Dauer, 1983; Major, 1981; Pointon, 1980). The problem is, if perfect isochrony can not be predicted, then just how much isochrony is required to be taken as support for hypotheses about stress timing vs. syllable timing?

One issue that reopened the debate about the temporal basis for speech rhythm was encountered in the late 1970's with the investigation of so-called perceptual centers or P-centers. This is basically a problem of measurement. As noted above, periodicity implies some regularly occurring events which may not be identical but only similar on successive cycles. However, the observable events that can be easily measured differ from syllable to syllable. A syllable can begin with just a vowel or a single consonant or a cluster of up to 2–3 consonants. Which point in one syllable should be lined up with (that is, treated as the same as) which point in another? From experiments in which subjects were asked to read a simple list isochronously, eg, "*ba, spa, ba, spa, . . .*", it became apparent that the time points that are adjusted to equal spacing may not have simple correspondence with any single physical feature of the signal. Although the experimentally observed P-centers appear to depend on several constituents of a syllable, a good first approximation appears to be the point of the onset of voicing (Morton et al., 1976). The models of Marcus (1981) and Scott (1993) base most of their success on the determination of an increase in energy in the spectral range of the first two formants. It should be noticed, however, that the deviations from isochrony noted in the P-center literature are not sufficient to account for the lack of observed isochrony. The problem of perceived timing appears to be more far-reaching than this.

It appears that listeners impose regularity on the speech signal that reflects their ability to predict what will happen and when. Success at these predictions gives rise to a strong experience of periodicity. We suspect that the grammar used by speakers incorporates an oscillatory system that generates rhythmic structure during speech production and also internally generates a similar perceptual rhythm when listening to speech (cf. Jones and Yee, 1993, on metrically based expectancies when listening to music). Thus, we agree with Abercrombie that listeners "have an immediate and intuitive apprehension of speech rhythm." Our goal is to discover what this apprehension might consist of. We will propose below a specific class of mechanisms that can begin account for these intuitions.

But first, we should point out that despite the great difficulty in finding regularly spaced time intervals in spoken English, there exists at least one well-attested case of highly regular timing in the production of speech—the case of Japanese, which has been shown to exhibit a simply defined periodicity. Thus if there are universals of linguistic rhythm, we should not be surprised if other languages exhibit periodicities of some sort as well.

## 2 Regular Timing in Prose: Japanese Mora

Japanese has traditionally been described as exhibiting 'mora timing' (Bloch, 1950; Han, 1962). In fact, one point of traditional language pedagogy that is taught to Japanese school children is that "all moras (Jap. *onsetsu*) have the same duration." In the archetypal case, a mora is a CV syllable. Thus, for instance, *doko* has two syllables and counts as two moras. But the first consonant in a cluster and the second part of a long vowel are also considered moras on their own. Thus, the $n$ in *Honda* and the first $t$ in *chotto* (even though there is no acoustic event separating the two $t$s) should have mora duration as well, making both words 3 moras long, although they have only two syllables. The city name *Tookyoo* (or *Tōkyō*) has 4 moras and two syllables. These traditional assertions about mora duration have now been verified—but only as long as the claim is interpreted in just the right way. The most obvious experimental hypothesis derivable from "all moras have the same duration" might be that the acoustic segment boundaries separating moras will be equally spaced. Thus, the interval from the onset of the *n* in *Honda* to the onset of the *d* should be the same as the duration of the acoustic *h* plus the *o* and the same as the *d* plus following *a*. Unfortunately, these intervals are all found to be quite different in duration.

A more sophisticated theory of the mora (first made explicit in Port, Al-Ani and Maeda, 1980) works much better and has been confirmed by later results (e.g., Port, Dalby and O'Dell, 1987; Han, 1994). This hypothesis about the mora asserts that, rather than look at a single mora at a time, one should examine only longer stretches of speech that contain several moras. In this way, one finds a strong tendency for whole words, for example, to come in durations of $nD$, where $D$ is the mean mora duration. The significance of this small change in formulation of the hypothesis is that individual moras may deviate quite strongly from the mean mora duration of an utterance, whenever the mora includes consonants or vowels whose intrinsic durations are particularly long or short or when a mora is made up of just a single C or V segment. Due to compensation in the durations of adjacent moras, however, each mora still contributes the same amount of duration to the duration of the word or phrase. This observation rationalizes the auditory intuitions about regular mora timing of Japanese speakers, but it raises difficult questions about the kind of mechanism that could underlie speech production. It also poses a challenge for mechanisms for perception that can account for these regularities that are apparent to native speakers of Japanese (and even to some second-language speakers, cf. Bloch, 1950), at least after being called to their attention by teachers.

To make sure the empirical phenomenon is appreciated, let us look at some Japanese timing data. In one experiment, several sets of words were constructed by lengthening words by one mora at a time (Port et al., 1987). As shown in Table 1, this yielded a number of words at each word length from 1 mora to 6 or 7 moras. Then the 28 words were embedded in a short carrier sentence and spoken by 5 native Japanese talkers at a self-selected comfortable speaking tempo and again at a self-selected faster rate. The whole duration of each word was then measured. Figure 1 shows the duration of the words in each of the 5 series for each word length averaged across the speakers. This duration mean is plotted against the number of moras in each word, from 1 to 7. The left panel shows the word durations at the slower speaking rate and the right panel shows the faster rate. Two simple facts can be seen in each panel. First, all words with the same number of moras have very nearly the same mean duration (at a given speaking rate). Second, the addition of one mora to a word lengthens the word duration by a nearly constant amount (assuming a constant speaking rate). That is, each mora, no matter what its segmental content seems to add the same amount of duration to the word. Finally, comparison of the left and right panels shows that changing speaking rate is primarily a matter of changing the mean duration of the mora.

| No of Moras | Word | English gloss | Word | English gloss |
|---|---|---|---|---|
|  | KA set | | HI set | |
| 1 | ka | scent | hi | sun |
| 2 | ka'ku | write | hika' | subcutaneous, hypodermic |
| 3 | kakusi' | pocket | ki'kka | *faux pas*, misstatement |
| 4 | kakusi'do | trapdoor | hikka'ku | to scratch, claw |
| 5 | kakusigoto' | secret | hikkake'ru | to hang, hook |
| 6 | kakusido'koro | hiding place | kikkakena'i | not to hang |
| 7 | | | hikkakerare'ru | get hung |

Table 1: Two of the five word sets used in Port et al. (1987), Experiment 2. The incremental word sets ranged from 1 to 6 or 7 moras. Each word was embedded in a simple carrier sentence and pronounced 4 times each at normal and fast speaking tempos by 5 native Japanese talkers. The total duration of each word was measured.

So, the mora is very regular, just as the intuitive tradition had it. But to see this simple regularity, one must measure durations in just the right way. One cannot measure a mora in isolation, but only in the context of measurements of its neighboring moras. Isolated measurement will lead to the (correct but misleading) observation that individual moras differ greatly in their duration (Beckman, 1982). Let us look more closely at the internal timing of a few words. Figure 2 shows the segmental durations of another set of words produced in carrier sentences (Port et al., 1987, Experiment 3). The words are *kuka, kuga, kaka* and *kaga* (some of which are nonsense words in Japanese). One sees in the figure such well-known (and putatively universal) segmental-timing effects as that "low vowels are longer than high vowels" and "voiceless consonants are longer than voiced consonants". If one measures one mora at a time, one obviously cannot find support for the claim that "moras are all the same duration". Thus, for example, *ka* is consistently longer than *ku* (in the same context). On the other hand, the duration of the words they appear in differ by much less than the difference in vowel duration. This is because other segments within the mora and the adjacent moras, *ka* and *ga*, have adjusted their duration to compensate. The main effects here, including the effect of medial consonant voicing on the initial stop, are all statistically significant (Port et al., 1987).

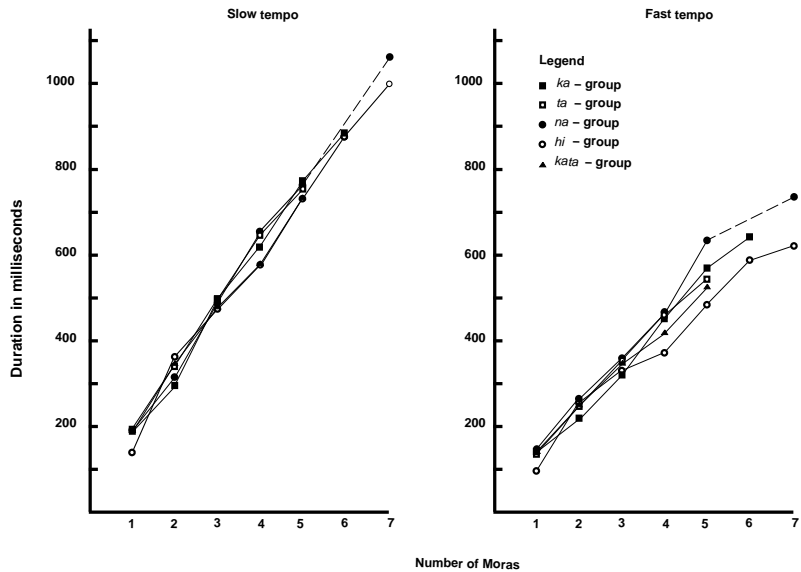

Figure 1: Results from Port et al. (1987), Experiment 2. The plot shows word duration as a function of the number of moras at two different speaking rates. Both plots are highly linear in character, showing that each mora causes the word duration to increase by a fixed amount at a given tempo.

Japanese has other segmental phonological rules, such as the one that devoices *i* and *u* between voiceless consonants, making phonemic *sukiyaki* be pronounced more like *s:kiyaki* (See legend in Figure 2). But this rule and others do not interfere with the moraic temporal structure of a word (Beckman, 1984). Not even word boundaries (Port et al., 1987) or imposition of contrastive stress (Bradlow et al., 1995) perturb the mora-timing constraint. Some lengthening of moras before phrase boundaries in conversational speech, however, is now well-attested (Takeda et al., 1986).

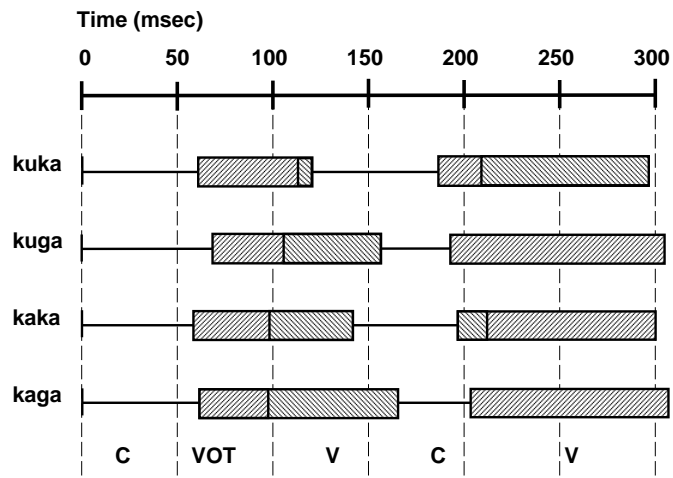

Figure 2: Results of Experiment 3 of Port, et al., 1987. Segmental durations for 4 mora Japanese words averaged over 6 repetitions by 9 Japanese speakers. For each word, the horizontal line to the left represents the duration of the initial /k/; the two filled boxes following represent the voice-onset time and the voiced vowel duration. The next line is the medial stop closure interval, followed by boxes representing voice-onset time and the final /a/. Note that the first vowel in the word /kuka/ is largely devoiced and thus is classified as aspiration.

These results show that the traditional observation about the regularity of mora timing is remarkably accurate, but *only* when time is measured in some way that smoothes the measurements over a time window that includes at least several moras. Compensatory durational adjustments assure that moras are generated at a regular rate overall, even though isochrony of individual moras is *not* observed. We need to inquire what this implies about the actual mechanisms that Japanese speakers and hearers employ to produce and perceive Japanese. Looking more broadly across languages, any success at finding a simple regularity in a single language suggests the possibility that other languages too may have highly constrained rhythm, but that, again, we may need more appropriate methods of measurement to find it—methods that are driven by input events extracted from the speech signal and which are capable of smoothing out the predicted rate of presentation over several periods.

## 3  Oscillator Entrainment as a Model for Speech Rhythm

One way in which the brain could perceive such quasi-periodic durational patterns as being regular when they are only partly so is if the linguistic perceptual system employed an adaptive oscillator (AO) resembling the model developed by Devin McAuley in our lab (McAuley, 1995b; McAuley, 1994). After describing the general model, we will suggest how this type of oscillator can help us understand how Japanese listeners deal with the problem of measuring the duration of a vowel in a difficult perceptual task, and then how it might be extended to develop a general theory of meter for language and music.

**The Adaptive Oscillator**  This general conceptual device, proposed by McAuley (1994, 1995b), is a system that generates a sinusoidal activation function at some particular frequency, as shown in Figure 3A. As in a similar model for musical meter recognition by Large and Kolen (1994), the instantaneous activation state is changing at all times and the rate of change is subject to modification by external input signals. In McAuley's model, inputs are constrained to take the form of pulses, shown in Figure 3B (rather than, for example, assuming continuous coupling for all points in time be-tween the input and the AO).[2] Coupling by discrete reset means that each time a strong input pulse occurs, the AO resets its phase to zero and immediately restarts its cosine function, as shown in Figure 3B. If the external pulses happen to be periodic, or *close enough* to periodic, then the AO responds by adjusting its intrinsic period to match the input period more closely, as shown in Figure 3C. System parameters determine the tolerance of the oscillator for variation in the period of the input, and the number of cycles needed for adaptation to be nearly complete. A modest amount of variation will not prevent entrainment to a moving average of the period of the input, and may even improve entrainment in certain circumstances (see McAuley, 1995b).

This system can be implemented computationally with just a small set of equations. It will quickly (within a few cycles) speed up or slow down so as to approximate the frequency of the inputs. If the inputs cease, then after a few cycles the system begins to decay toward its original intrinsic rate. Simulations show that if the inputs are somewhat irregular or if one or two pulses are missing entirely, the behavior of the AO is largely unaffected; it oscillates at a rate that approximates a running average of the most recent input cycles (as shown in Figure 3D).

McAuley developed this model to account for human listener performance on a task of discriminating changes in the rate of a series of short tones (McAuley, 1995a; Drake and Botte, 1993). In each trial of this task, subjects first heard a series of beeps (with the series varying in length from 2-7 beeps), then a pause, then another beep series. The subjects had to indicate if the second series was faster than the first. McAuley showed that many details of subject performance on the tone series with periods ranging from about a tenth of a second to a full second could be accounted for on the hypothesis that they were making their discriminative judgments by employing an internal mechanism that closely resembles an adaptive oscillator. To account for the data, the intrinsic period of the AO needed to be set at around a half second. His program was presented with real-time stimuli and performed the same discrimination task as the human subjects.

**Oscillator Theory of Mora Perception.** It is not difficult to imagine that such an adaptive oscillator could provide the perceptual basis for the regularity of mora timing in Japanese. Thus, if the AO finds itself able to entrain easily to a pulse series, then listeners should tend to report high temporal regularity. The production measurements above show that when Japanese speak, they somehow compensate for intrinsically short or long moras by both 'anticipatory' and 'perseverative' lengthening or shortening in adjacent moras. The effect of these short-term adjustments in speaking rate is to keep the average mora duration constant. All that is required to apply an AO to the perception of Japanese is an appropriate method of converting speech into onset pulses—into something resembling P-centers.

---

[2]Continuous coupling is the type more commonly addressed in physics textbooks. One simple mechanical model of continuous coupling would be two different pendula connected by a bungee cord. The bungee cord assures that the instantaneous amplitude of each affects the amplitude of the other at all points in time.

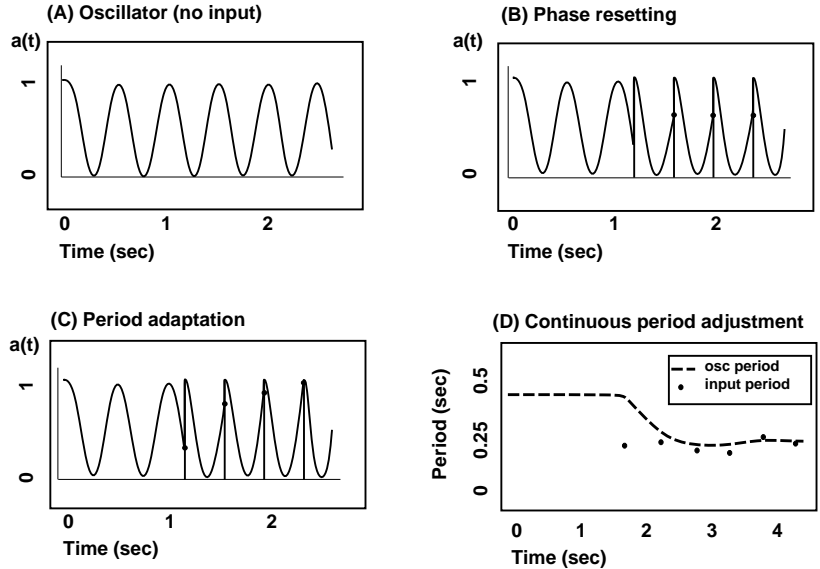

Figure 3: (A) The sinusoidal activation function of an oscillator in the absence of external input. It is convenient to raise the cosine to the range [0,1]. (B) Periodic input pulses begin here just after 1 sec and are added to the oscillator's activation. In the figure, the pulses are superimposed on the activation. When the sum exceeds 1.0, the oscillator is phase reset to phase 0. (C) Each time the oscillator is phase reset, it adjusts its period to be more like that of the input, resulting in gradual synchrony, or entrainment, between oscillator and input. (D) Oscillator periods (dashed line) and input periods (dots) are shown over time. Variability in the input pulse periods tends to be ignored or smoothed out in the oscillator's running estimate of the input period. When input ceases, the oscillator will slowly decay back to its intrinsic period (not shown).

If there is an adaptive oscillating mechanism that is excited by moraically produced Japanese speech, then we might expect that Japanese listeners could use the phase angle within the mora as a time measure in making perceptual judgments. They might obtain that phase angle measurement from certain physical properties of the moras preceding the syllable being listened to.

Is there any data to support such a perceptual model? Certainly there is considerable data on English showing that listeners make use of timing in the context of a word to make perceptual judgments (Port and Dalby, 1982; Dorman et al., 1979). So there is good reason to expect that Japanese listeners also use temporal details of speech to adapt their perceptual processes. The AO model generates many specific predictions which can be tested. But our hypothesis is that it is the *grammar of Japanese* that defines this mora-based oscillatory system. Under this hypothesis, the distinctive temporal structure of the language, which constrains both speakers' productions and their perceptual systems, is assigned to the grammar of the language itself.

Although there are many experiments waiting to be done to explore implications of this model, enough has been learned about Japanese to show that a quite simple and regular temporal structure can be found for this language. And an adaptive oscillator is a plausible first-order model of the internal clock employed by Japanese listeners for measurement of speech timing. Of course, such a revision of the formal nature of grammar will necessarily have wide-ranging implications for linguistic theory. Those will have to be explored at a later time.

But to return to the earlier problem of English stress timing, can any analogous mechanism be proposed for English? If English does have any 'stress-timed' characteristics, then the implication is that there must be oscillators nested at more than one time scale. That is, the hierarchical relationship of syllables and feet implies that there would have to be a syllable-level oscillator coupled to a foot-level oscillator that measures the inter-stress interval. This is a simple case of what is often called *meter*—a hierarchical system of cycles. Within linguistics, meter is usually defined in terms of an ordinal time scale—typically as ordered strings of Strong and Weak units at several hierarchical levels. But in music, meter presumes a ratio scale for time. The evidence presented above shows that Japanese does employ a ratio scale for time. Could English stress timing be interpreted in ratio-scale terms? It is possible that previous failures in this regard reflect the use of inadequate hypotheses about the perceptual mechanisms involved.

In the following sections we shall investigate English speech rhythm, by looking for evidence that speech production is rhythmically and metrically constrained. Note that when looking at dynamics, we can no longer retain the traditional stance of linguistics research by addressing only 'la langue' or 'linguistic competence', or any sort of idealized, style-free speech. Since we can only do research on actual speech performance in time, we must choose some particular style of speech. The style that is most useful will depend on our purposes. Of course, any actual speech style—whether the 'public narrative' style (eg, telling a story to a group), 'intimate conversation', a 'public declamatory' style (eg, preaching), or poetry performance—will necessarily be influenced by nonlinguistic factors as well. This cannot be helped, and presumably our model of the dynamics of the language will be compatible with any style of speech performance, as long as the performers are speaking the language. If Abercrombie was right when he claimed that "the rhythm of everyday speech is the rhythm of verse" (1967, p. 98), then it is likely that the style of speech found in a phrase repetition task would encourage language-specific temporal structure to be rhythmically streamlined. We might expect that speakers could be forced into certain 'attractors'. If this happens it would simplify our modelling task by allowing us to direct our attention to a speaking style that has greater regularity than might be found in, say, a public narrative style.

## 4 English rhythm in a phrase repetition task

Many observers seem to hear English stressed syllables at perceptually equal intervals. This suggests that the appropriate level at which to look for rhythm in English might be the organization of stresses (and hence feet) within a phrase. Such a hierarchy, if it were to be completely regular, would resemble the metrical hierarchies of music, which tend universally to be organized around a fixed number of *beats* (most often 2, 3 or 4) per *measure*.

Given what we currently know about the details of Japanese timing, such nested hierarchical structures are not obvious (but bimoraic feet have been proposed for this language based on phonological evidence (Poser, 1990).) But how can we discover the extent to which English has a music-like hierarchical rhythmic structure?

One method would be to ask speakers to read a poem or song lyric. But such tasks impose hierarchical timing constraints that are artistic inventions at which speakers may be expected to differ greatly in their performance skill. Instead we used the simple phrase-repetition task, a method with a tradition going back to Stetson (1951). This involves asking speakers to repeat a phrase over and over in succession, once for each metronome pulse. In our experiments, subjects were asked to repeat a phrase with prescribed timing for a stressed syllable within the phrase. If English is rhythmical at the level of stressed syllables, then these phonological interstress periods should interact with any other period that might be imposed on the performance—such as the period of the repetition of the whole phrase.

To see why this should be expected, consider other motor tasks in which subjects are asked to maintain two oscillators at frequency ratios that are complex—that is, at ratios that are not 1:1, 2:1 or 3:1 (Kelso and de Guzman, 1988; Treffner and Turvey, 1993). In one variant of Treffner and Turvey's task, for example, seated subjects were asked to swing a pendulum with the right arm and wave a drumstick with the other arm at prescribed rates. They were told simply to maintain each rate independent of the other. However, subjects could not avoid allowing the two arms to couple in such a way that the actual periods of the two arms had a strong tendency to fall into certain simple integral relationships to each other. Thus if the ratios of the prescribed rates for the pendula were 3:4, subjects had a strong tendency to slip the speed of one or the other in such a way that the ratio would become either 1:2 or 1:1. These ratios were interpreted as attractors of the motor control system.

Our experimental task resembles these tasks by calling for a similar conflict between an imposed phrase-repetition rate (supplied by a metronome-like stimulus) and the hypothetical phrase-level stress periodicity of English. Thus, we expected that subjects' productions would tend toward simple integer ratios of the inter-stress interval to the phrase as a whole. If this occurs, then we should find prominent events associated with the stressed syllables located at or near favored phase angles of 1/2, 1/3, 2/3, etc. We expect this for much the same reason that prominent musical events tend to occur at harmonic fractions of the musical measure. Thus the nominal phase of the onset of the third note in Figure 4 A, which is in $\frac{3}{4}$ time is 0.666, while the same note in Figure 4 B, which is in $\frac{4}{4}$ time has a nominal phase of 0.5. Measured phases may deviate somewhat from these nominal values as a result of expressive timing (Clarke, 1989), systematic bias (Church and Broadbent, 1990) or inherent variability (Wing et al., 1989).

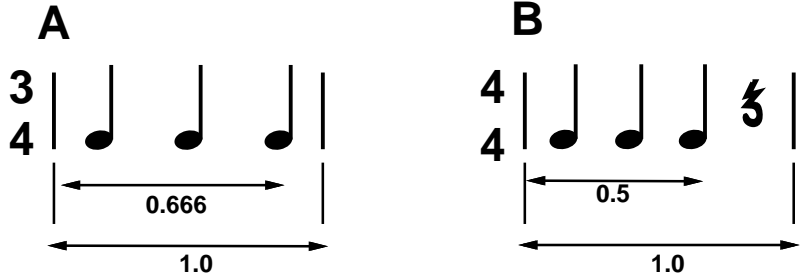

Figure 4: How rhythmic structure can be inferred from phase. In A, the measure (phrase) has a 3 beat (foot) duration and the third note (stress) begins at a phase angle of 0.666. In B, the same note begins at a phase of 0.5, because the measure lasts for 4 beats.

The goal of this experiment was to see whether, in repeating a phrase over and over, English speaking subjects would show any bias for placing highly salient phonetic events, such as syllable onsets, at phase angles that imply rhythmic or harmonic fractions of the total duration of the phrase. Thus, a task was devised that called for them to repeat a phrase and locate a stressed syllable onset at an arbitrary phase angle with respect to the repetition rate of the phrase. Our expectation is that this task will prove very difficult and subjects will be biased toward certain harmonically related phases.

**Method.** For both stimuli and human production data in the following experiment, we needed to provide a definition of 'the moment of occurrence' or beat, of a syllable. This should be in approximate correspondence with the time point in the syllable which a subject would use for lining the syllable up with another stimulus and should thus be close to the syllable's P-center. We used an algorithmic definition of a syllable's acoustic "beat" (based in part on the insights of Scott, 1993, and Marcus, 1981).

After recording the speech at 8000 samples per second directly onto disk, it was passed through a simple auditory filtering model using the Lutear software package. The signals were passed through a bank of 6 gammatone filters distributed over the range 300 to 2000 Hz. This range preserves most energy from the first two formants while F0 and high frequency frication were largely filtered out. The 6 resulting filter outputs were summed and smoothed to yield an estimate of total signal energy in this range. The signal was rectified and smoothed again. This yielded smooth amplitude contours like the one shown in Fig 5. This might be interpreted as a continuous measure of 'sonority'. Any rise in this amplitude can be associated with an acoustic beat. The moment of occurrence of the beat was defined as the point halfway between the local maximum and the preceding local minimum. Its magnitude was defined to be proportional to the size of the associated rise in amplitude and scaled to the range $(0, 1)$. This is what is meant by a syllable's "beat" in what follows. With this algorithm, beats are usually located very near the vowel onset of a syllable. Obviously, the optimal definition of such a beat is an empirical issue. The beat provides a localized event which can serve as resetting trigger for an entraining oscillator.

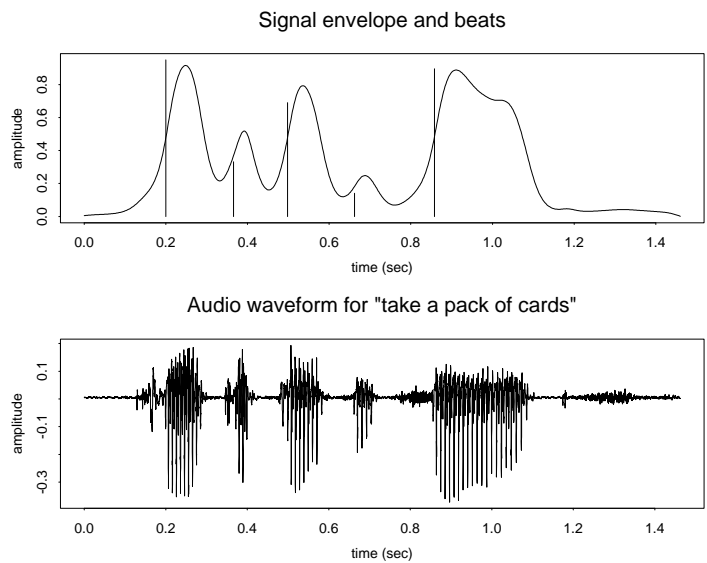

Figure 5: Beats associated with one utterance of the phrase *Take a pack of cards*. The top panel shows the smoothed amplitude envelope as described in the text. Extracted beat amplitudes are overlaid. The waveform is shown in the lower panel. It can be seen that the beats lie close to the vowel onsets.

Stimuli were constructed that consisted of 8 repetitions of the pair of words *take* and *cards*, produced consecutively and with various spacings. The individual words of the stimuli were generated using commercial speech synthesis software (from Eloquent Technologies) using a flat intonation contour. The interval between the beats (as defined above) of successive tokens of *take* was fixed at 1.5 seconds, and this served as a complete cycle for defining phase measurements. The phase angle at which the beat of *cards* occurred was varied across the trials. This will be referred to as $\Phi_{cards}$, and we use the convention of specifying phase in the range [0,1]. Eight values of $\Phi_{cards}$ were used, ranging from 0.3 to 0.65 in increments of 0.05.

On each trial, subjects were presented with the stimulus sequence over headphones. Their task was to try to repeat the phrase *take a pack of cards* in time with the stimulus, such that their productions of *take* and *cards* were temporally aligned with the stimulus. They were asked to begin talking on the second repetition of the stimulus (i.e., the second time they heard the word *take*), and to continue speaking after the stimulus stopped. When they had completed 7 repetitions of the phrase without accompanying stimulus, they were signaled to pause for 3 seconds, and then to produce 7 more repetitions, again trying to maintain the relative timing of *take* and *cards* given previously by the stimulus. Subjects were not required to count the number of repetitions they produced, but they received visual signals from the experimenter indicating when they should pause, and when they should stop. The data on each trial thus fell into three groups—tokens produced simultaneous with the synthetic stimulus words, those produced immediately after the stimulus sequence ceased, and those produced after the pause. There were 7 tokens in each group, giving 21 tokens per trial. There were three blocks of trials per subject; one in which the target value of $\Phi_{cards}$ increased from 0.3 to 0.65 across trials, one in which it decreased, and one in which the target values of $\Phi_{cards}$ were randomized. Subjects were given a short break between blocks to avoid fatigue. A total of 6 subjects was run, allowing the order in which the blocks were presented to be counterbalanced across subjects. All subjects were students outside of the field of linguistics, and were naive to the purpose of the experiment. The results presented here represent an analysis of the data from 4 subjects.

**Data analysis.** From each group of 7 productions of the target phrase, the first and last were discarded, and the value of $\Phi_{cards}$ for the remaining 5 productions was measured using the beat extraction process described above. The phase at which the beat of *cards* occurred was measured relative to the preceding and following beats associated with the subject's productions of *take*. There were 5 phase measurements per group with 3 groups per trial.

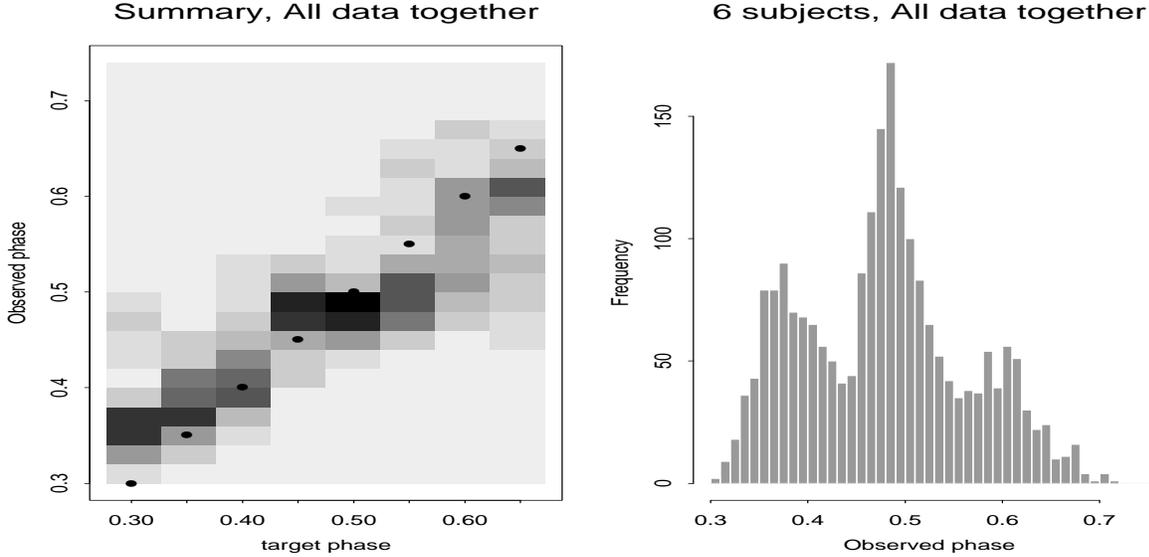

Figure 6: Summary of all $\Phi_{cards}$ values for all 6 subjects together. Left panel: Target phase versus observed phase across trials. The dots indicate the target phase and the darkness of the cell represents relative frequency of occurrence. Right panel: Cumulative histogram of all productions for all subjects. Note the presence of three broad modes, centered near 0.36, 0.49 and 0.6.

**Results.** We were interested in whether subjects could match the target phase or whether, as suspected, their productions would show a preference for locating beats at phases that represent harmonics of the phrase period, that is phases near 1/3, 1/2, 2/3, etc. Since the three within-trial data sets looked very similar, we report them pooled together in Figure 6. This figure shows results for all subjects together, collapsed across blocks and within-trial groups. From the left panel it can be seen that subjects' productions often have phase values quite different from those of the target phase (shown here with dots). Indeed, rather than approximating the straight line indi-cated by the dots, the data are bunched into a steplike function. This is confirmed by the right hand panel, which shows a cumulative histogram of all data for all 6 subjects. There are three clear modes, centered roughly at 0.36, 0.49 and 0.6.

The observed modes are close to those which would be predicted by a very simple musical representation of rhythmic patterns. If the phrase were repeated with the stresses of *take* and *cards* falling on the first two beats of a $\frac{3}{4}$ meter, we would predict a value of $\Phi_{cards}$ of 0.33. In either a $\frac{2}{4}$ or a $\frac{4}{4}$ rhythm, we would expect 0.5 to predominate, while a $\frac{3}{4}$ rhythm in which each of *take*, *pack* and *cards* fell on a beat, would give a phase of 0.66 for $\Phi_{cards}$. This result suggests that speakers have great difficulty placing stress at an arbitrary phase angle when repeating a phrase over and over. Instead, there are three highly attractive phases that are located at or near the lower harmonics of the period defined by the repetition of the entire phrase. This is strong *prima facie* evidence for a dynamic process of entrainment, rather than mere concatenation of individual durations.

Thus, the results are strongly suggestive of a hierarchical structuring of the timing of these repeated phrases. Just as in musical timing, subjects show a preference for a measure-like unit—the phrase as a whole— and a beat-like unit—the stressed syllable onset. Given that such preferences are so easily observed, it seems likely that even normal speech production may exhibit somewhat similar preferences.

# 5 Oscillator Models of Meter

How might a perceptual system lock onto the kind of hierarchically structured rhythm patterns that are observed here as well as in typical musical structures? A computational model of the perception of rhythm in music or speech would require the capacity to find meter in a sequence of input pulses. In this section, we offer some general considerations on the form of a hierarchical metrical perception system—a device suitable for recognizing metrically structured speech (or music). Finding meter in turn consists in identifying periodicity at two or more levels and locating the most prominent pulse (the 'downbeat') at each of these levels. The rhythm of both music and speech is characterized by rate invariance (Port, 1986; Port, 1990) and by some deviations from strict periodicity. Thus, a model must generalize across a range of tempos and be relatively insensitive to minor fluctuations. At the same time, a model needs to be constrained so that it does not find meter where there is none. In particular, it should 'expect' the downbeats at the different metrical levels to be lined up and the period at each level to be an integral multiple of those at levels below it.

Current oscillator models of periodic phenomena (McAuley, 1995b; Large and Kolen, 1994; Miller et al., 1988) begin with the periodicity built directly into the model in the form of one or more oscillators, simple processing units characterized by a period and an instantaneous phase angle. In all of them, the central idea is that oscillators with particular periods respond to phenomena in the world with similar periods. When an oscillator is activated, it can make predictions about when prominent events are to occur and use these predictions to increase sensitivity to events at particular points in time.

An *adaptive* oscillator (Large and Kolen, 1994; McAuley, 1995b) has the capacity to entrain to periodic phenomena by synchronizing its period with the periods inherent in the stimulation and aligning its zero phase with the strongly accented events in the input patterns. Adaptive oscillators entrain to an input pattern through two sorts of *coupling*, phase and period coupling. As we have seen, phase coupling with a sequence of pulses consists in adjusting the phase of the oscillator so as to lessen the distance between an input pulse and the oscillator's zero phase. Since the period of the input pulses may be quite different from that of the oscillator, the oscillator should not respond in an unconstrained fashion to all pulses. For this reason, phase coupling occurs only within a window around the oscillator's zero phase. The degree of phase adjustment depends on the particular model. There are two main variants, phase-reset models and continuous coupling models. In phase-reset models, like McAuley's model described above, the oscillator resets its phase to zero each time an input pulse occurs within its coupling window. In continuous models (e.g., Large and Kolen, 1994), the phase adjustment is a function of the deviation of the oscillator's phase from zero when the input pulse occurs.

Period coupling (that is, frequency adjustment) is required for smooth performance if oscillators may begin with periods different from that of the input signal. Period coupling is similar to phase coupling except that there is no analog to phase resetting (since the mere lack of period match does not indicate which direction to change the frequency). Oscillators also need a *resting period* toward which their period decays. Without this decay, the period of an oscillator can increase or decrease in an unconstrained fashion in response to inputs.

Once an adaptive oscillator has "found" the input period and synchro-nized its zero phase with the input pulses, it remains entrained to the input as long as the input remains periodic. Minor deviations in the period of the input or slow changes in rate are accommodated by the oscillator as it adjusts its phase and period to the pulses. Even a small number of missing pulses will not throw off the oscillator because it will continue to oscillate at roughly the expected period for short intervals. It is the capacity to lock onto periodic input patterns and the resilience with respect to noise and missing inputs that give oscillator models their power.

In order to work toward development of a hierarchical system appropriate for meter, an oscillator also requires an associated activation or output that will serve as a local measure of how well it matches either external input or the rate of other oscillators. In McAuley's model (McAuley, 1995b), output is a weighted sum of the oscillator's phases when it is reset. Thus, it is a useful measure of how well the oscillator is synchronized to recent inputs. In Miller, Scarborough, and Jones' music model (Miller et al., 1988), oscillator activation measures how well an oscillator characterizes the input pattern and how well it agrees with other oscillators in terms of a set of metrical constraints. In their model, oscillators are activated when their zero phases coincide with input pulses and activated or inhibited by other oscillators on the basis of constraints built into the network.

As for discovery of hierarchical metrical structure, the Miller, Scarborough and Jones model showed how inter-oscillator activation and inhibition can aid a network of oscillators (although non-adaptive ones) in finding the meter in musical input which has its micropulse level artificially synchronized with the oscillators' periods. The musical meter identification task of this model is somewhat similar to our own task of modeling the perception of speech rhythm, but differs in the several crucial ways. First, the metrical structure of speech is considerably more elusive than that of music, and exhibits more irregularity at a given metrical level. In addition, a particular meter is likely to apply to only a short stretch of input (in normal speech). A model suitable for speech must have the capacity to discover meter very rapidly and make full use of the constraints among different metrical levels in order to do so. Second, our goal is the identification of particular rhythmical patterns, so the discovery of metrical structure serves this purpose, but is not an end in itself. A model suitable for speech needs a way of *representing* rhythmical patterns, something which is not provided by a set of unconnected adaptive oscillators or by the simple connections between the units of Miller, Scarborough, and Jones' network (Miller et al., 1988).

We are working on development of a model that consists of a network of adaptive oscillators with varying resting periods and one or more input units that are excited by pulses of varying amplitude. Each oscillator should behave like a unit in a connectionist network in that it has an amplitude and is joined to other oscillators by weighted connections. Unlike other adaptive oscillator models, here the oscillators should couple with each other in a way that facilitates the identification of hierarchical metrical structure and also permits the representation of rhythmical patterns that fit that meter. The inter-oscillator connections will need to include both built-in components and components that are learned as the network is exposed to input patterns. Unlike the connectionist model of Church and Broadbent (1990), the coupling strengths and the oscillator periods will be adaptive variables in the model.

The constraints on a system capable of learning metrical patterns turn out to be quite subtle. However, all but the learning component of the model have already been implemented in at least preliminary versions.

# 6 Discussion

This work described here has implications of several sorts. First, by proposing dynamical models as intrinsic components of the phonology of a language, we are implicitly suggesting radical changes in the nature of phonological theory and linguistic theory as a whole. On the other hand, by providing a description of language in dynamic terms, this approach creates the possibility of unification of the theory of language with theories of motor control and perception. That is, linguistics from this point of view may be integrated once again with the rest of cognitive science (see Port and Van Gelder, 1995) where dynamical modeling is becoming widespread to the point of becoming almost the default framework for cognitive modelling.

The type of scientific explanation offered by a dynamic model is greatly preferable to explanations given by symbolic descriptions because dynamic models work in real time. In contrast, symbolic models always require some external system (the 'analog-digital converter' or 'phonetic implementation system') to translate into and out of real time. Because of this property, dynamical models resemble the explanations offered by physicists for many morphological structures that have nothing at all to do with speech. For example, periodic structures in time and space are widely found in the natural world: uniform tubes and strings resonate at harmonically related frequencies, crickets chirp at a regular rates, clouds and sand dunes are of-ten arranged in spatially periodic 'streets', tigers, zebras and butterflies grow regular periodicities of coloring, fireflies and cicadas sometimes entrain themselves to each other in time. Mathematical accounts of these phenomena appeal to the behavior of relatively simple dynamic systems operating in the atmosphere, in embryonic zebra skin, in the firefly brain, and so forth (Haken, 1983; Kelso et al., 1994; Murray, 1993). It is not implausible that very similar dynamic models apply to the mechanical, neural and cognitive mechanisms for language. One consequence is that speech in many languages may exhibit periodic temporal structures, and such patterns are perceived by human listeners (at least by listeners with appropriate experience in the language) as exhibiting periodic temporal structures even though the regularities may be far from obvious in visual displays.

Of course, if such periodicities are natural aspects of the physical world, perhaps the rhythm of speech is simply a mechanical universal of the human body and thus has nothing whatever to do with linguistic cognition. One might argue that it arises simply from the constraints on how a complex group of muscles needs to be coordinated, and that such coordination is most naturally handled by an oscillatory system. If this were the case, however, speech rhythm should be completely uniform across languages, and would be more akin to 'universals' of locomotive gaits or respiration. However, there is overwhelming evidence of differences from language to language in the way timing is controlled. The mora structure of Japanese is peculiar to this language, and the rhythmic structure of English is found only in English and its close relatives. Indeed, precisely these differences constitute a major source of foreign accent and are an important reason why foreign accented speech has reduced intelligibility (Tajima et al., 1994). Thus, these results support the hypothesis that the phonology of a language is a description of a motor and perceptual *skill* that may make opportunistic use of any mechanical or other constraints on coordination, but which is still obviously a cognitive function on par with other high level functions.

# 7 Conclusions

The results we have surveyed here in Japanese and in English justify the claim that speech is often rhythmical. In Japanese, native speakers' intuitions about their speech timing are shown to be well grounded in fact when a sequence of moras are considered, rather than single moras in isolation. The result of the production process used by Japanese speakers is a sequence of physical events which are very periodic when attended to by an appropriate perceptual mechanism. The adaptive oscillator, whose development was initially motivated by findings on the perception and estimation of simple temporal intervals, provides a model for a perceptual process which can 'regularize' the sequence of physical beats, producing a smoothed measure of temporal regularity. This would account for the strong perceptual impression of regularity in the face of variability in timing.

In English, we have begun a series of experiments designed to identify the rhythmic basis for speech production. Perhaps appropriate mechanisms can be found that will rationalize the perception of regularity here as well. In the early work on this project reported here, a single repeated phrase was shown to be uttered preferentially in one of three rhythmic forms, corresponding roughly to either a 3-beat or a 4-beat meter by speakers with no special training or instruction. Again, the physical signal is not perfectly regular, but a regularizing psychological model which can map from variable physical beats to a smoothed estimate of periodicity may account for perceived regularity.

We propose this work as the first steps toward establishment of a 'phonology of time,' a proposed subfield of linguistics concerned with the structure of speech and language in time. One goal is to provide an explanation for the observed temporal structure of speech in various languages in terms of 'natural law', both at the microstructural level of segmental variation, and the macrostructural level of speech rhythm and prosody. The resulting models will necessarily look quite different from conventional linguistic models. But the psychological mechanisms for adaptive oscillators are consistent with known psychophysical data and with known properties of nervous systems. And they can be expressed in terms that permit specification of real time rather than the pseudo-time of serially ordered strings of symbols.

**Acknowledgments.** The authors are grateful for helpful comments from J. Devin McAuley, Catherine Rogers, Kenneth de Jong, Keiichi Tajima and Stuart Davis. This research was supported in part by the Office of Naval Research, Grant number N00001491-J1261 and ATR Human Information Processing Research Laboratories (Kyoto).